\documentclass[conference]{IEEEtran}
\IEEEoverridecommandlockouts
\usepackage{cite}
\usepackage{amsmath,amssymb,amsfonts}

\usepackage{graphicx}
\usepackage{textcomp}
\usepackage{xcolor}
\usepackage{multirow}
\usepackage{amsmath}
\usepackage[utf8]{inputenc} 
\usepackage[T1]{fontenc}    
\usepackage{hyperref}       
\usepackage{url}            
\usepackage{booktabs}       
\usepackage{amsfonts}       
\usepackage{nicefrac}       
\usepackage{microtype}      
\usepackage{xcolor}         
\usepackage{colortbl}
\usepackage{graphicx}
\usepackage{amsmath}
\usepackage{algorithm}

\usepackage{algpseudocode}
\usepackage{adjustbox}
\usepackage{booktabs}
\usepackage{multirow}
\def\BibTeX{{\rm B\kern-.05em{\sc i\kern-.025em b}\kern-.08em
    T\kern-.1667em\lower.7ex\hbox{E}\kern-.125emX}}
\begin{document}
\title{FALCON: \textbf{F}eedback-driven \textbf{A}daptive \textbf{L}ong/short-term memory reinforced \textbf{C}oding \textbf{O}ptimizatio\textbf{N}}
\author{Zeyuan Li$^{*1}$, Yangfan He$^{*2}$, Lewei He$^{\dagger1}$, Jianhui Wang$^3$, Tianyu Shi$^4$ ,Bin Lei$^5$, Yuchen Li$^6$, Qiuwu Chen$^6$\\
    $^1$ School of Software, South China Normal University\\
    $^2$ University of Minnesota - Twin Cities \\
    $^3$ University of Electronic Science and Technology of China \\
    $^4$ University of Toronto \quad
    $^5$ University of Connecticut \quad
    $^6$ AI center, AIGCode Inc.\\
  \text{2023024326@m.scnu.edu.cn}\quad \text{he000577@umn.edu} \\
  \text{helewei@m.scnu.edu.cn}\quad
  \text{2022091605023@std.uestc.edu.cn}\\ 
  \thanks{$^{*}$ Equal Contribution.}
  \thanks{$^\dagger$ Corresponding Author.}
}

\maketitle

\begin{abstract}

Recently, large language models (LLMs) have achieved significant progress in automated code generation. Despite their strong instruction-following capabilities, these models frequently struggled to align with user intent in the coding scenario. In particular, they were hampered by datasets that lacked diversity and failed to address specialized tasks or edge cases. Furthermore, challenges in supervised fine-tuning (SFT) and reinforcement learning from human feedback (RLHF) led to failures in generating precise, human-intent-aligned code. To tackle these challenges and improve the code generation performance for automated programming systems, we propose \textbf{F}eedback-driven \textbf{A}daptive \textbf{L}ong/short-term memory reinforced \textbf{C}oding \textbf{O}ptimizatio\textbf{N} (i.e., FALCON). FALCON leverages long-term memory to retain and apply learned knowledge, short-term memory to incorporate immediate feedback, and meta-reinforcement learning with feedback rewards to address global-local bi-level optimization and enhance adaptability across diverse code generation tasks.Extensive experiments show that FALCON achieves state-of-the-art performance, outperforming other reinforcement learning methods by over 4.5\% on MBPP and 6.1\% on Humaneval, with the code publicly available.\url{https://anonymous.4open.science/r/FALCON-3B64/README.md}.
\end{abstract}
\begin{IEEEkeywords}
Code generation, Reinforcement Learning, Diverse Feedback.
\end{IEEEkeywords}
\begin{figure*}[t]
    \centering
    \includegraphics[width=\linewidth]{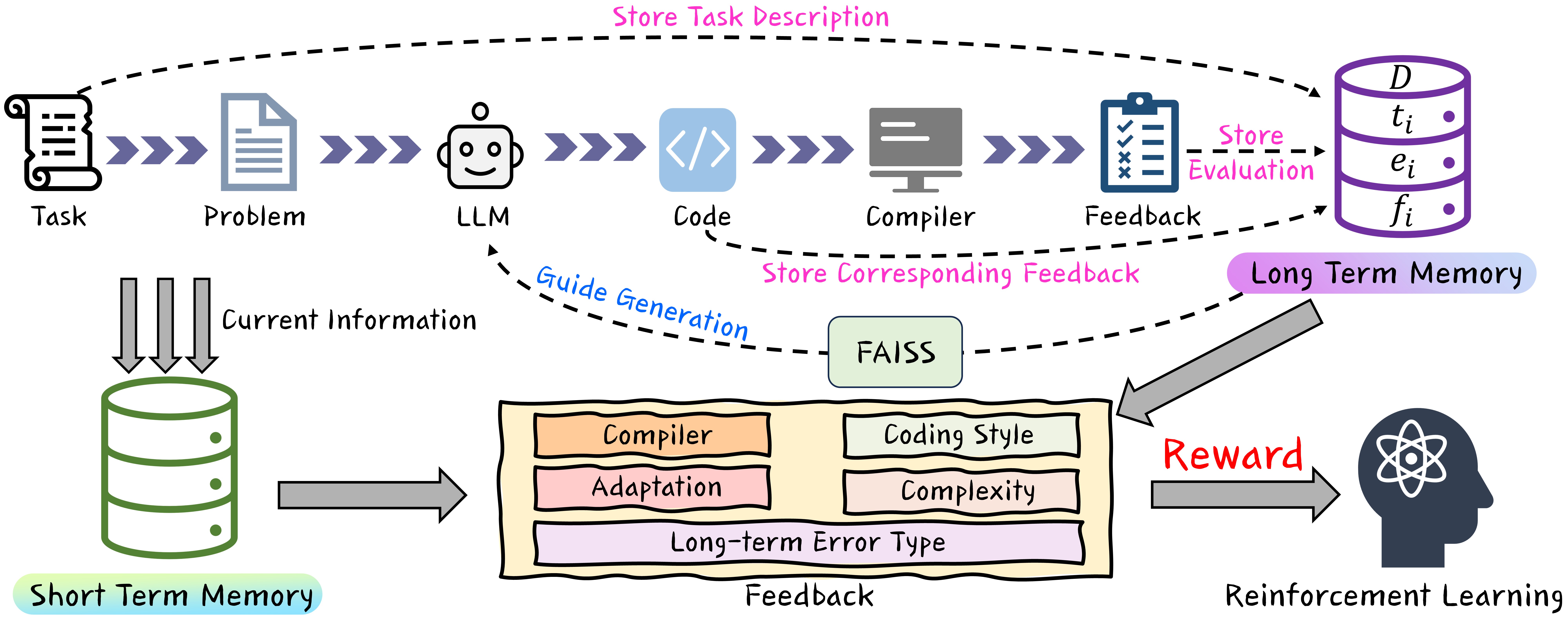}
    \caption{
The overview of the FALCON framework. Detailed feedback process. Reinforcement learning updates using feedback to compute RL loss and optimize the model (see Figure~\ref{fig:meta_framework} for the detailed meta-reinforcement learning framework).
}
    \label{fig:system_architecture}
\end{figure*}
\section{Introduction}
The development of Large Language Models (LLMs) has significantly advanced automated code generation \cite{LLMS:23}. Models like CodeLLaMA \cite{codellama:23} and DeepSeek-Coder \cite{deepseek:24}, tailored for code-centric tasks, have demonstrated outstanding performance across programming challenges. While LLMs excel in instruction-following through tuning \cite{jiang2024survey}, they often misalign with user intent, making feedback-based adjustments critical. For example, InstructGPT \cite{Instruct:22} leverages reinforcement learning with human feedback (RLHF), and CodeRL \cite{CodeRL:22} uses compilation feedback to refine model performance. Similarly, CompCoder \cite{CompCoder} enhances code compilability with compiler feedback, and RLTF \cite{RLTF:23} offers fine-grained feedback on compiler errors. However, current RL frameworks generate compilation errors and overlook non-differentiable features (e.g. coding style) that affect the performance significantly \cite{jiang2024survey}. To address these challenges, we propose a reinforcement learning system combining long-term and short-term memory feedback. From the global level, long-term memory tracks trends over time for higher-quality code retrieval, while from the local level, short-term memory captures recent errors and immediate feedback. In summary, the main contributions of this paper are as follows:
\begin{itemize}
    \item \textbf{Short-Term and Long-Term Memory for Reinforcement Learning}: We propose a dual-memory approach for reinforcement learning in code generation, where short-term memory enables real-time corrections and long-term memory accumulates knowledge from past runs to improve code quality and reduce repetitive mistakes.
    \item \textbf{Non-Differentiable Code Features into Feedback Loops}: Our approach addresses the limitation of current RL frameworks by integrating non-differentiable code features like style, readability, and best practices into the feedback loop, ensuring the generated code is both functionally sound and aligned with real-world programming standards.
    \item \textbf{Meta-Reinforcement Learning for Generalization Across Tasks}: We enhance the model’s versatility by incorporating meta-reinforcement learning, allowing it to efficiently generalize across diverse programming tasks, adapt quickly to new environments, and handle a wide range of coding challenges with fewer training iterations.
\end{itemize}
\section{Related Works}
\subsection{Pre-trained Models for Code Generation}
In recent years, pre-trained language models have made significant progress in the field of code generation. Trained on large-scale code corpora, these models have demonstrated powerful code generation capabilities. For example, CodeBERT \cite{CodeBERT:20}, a model based on an encoder-only architecture, has shown impressive results. With the advent of in-context learning methods, decoder-only Transformer models have become the dominant technology for language modeling \cite{AIAN:17}. Several models, such as CodeGeeX \cite{CodeGeeX:24} and DeepSeek-Coder \cite{deepseek:24}, use Causal Language Modelling (CLM) pretraining, while others like CodeT5 \cite{CodeT5:21} utilize an encoder-decoder architecture. Additionally, models like CodeFusion \cite{CodeFusion:23} leverage diffusion-based techniques for code generation. These pre-trained models exhibit great potential in code generation tasks, achieving notable improvements in accuracy through various architectures and training strategies. However, they still face challenges in ensuring the syntactical and functional correctness of the generated code.
\subsection{Reinforcement Learning on code}
Reinforcement learning (RL) learns optimal strategies through interaction with the environment and reward signals \cite{OPDR:19}, showing strong performance in sequence generation tasks. Code generation, however, is more complex than typical NLP tasks, as it requires not only syntactic correctness but also functional accuracy—ensuring the code compiles and behaves as expected. While unit tests are helpful for verifying correctness, they are insufficient unless well-designed to cover diverse and edge cases. To address this, methods like CodeRL \cite{CodeRL:22} incorporate unit test feedback into fine-tuning via reinforcement learning. PPOCoder \cite{EBCG:2023} improves this with Proximal Policy Optimization, and RLTF \cite{RLTF:23} further integrates error information and multi-granularity feedback. StepCoder \cite{StepCoder:24} enhances the process using compiler feedback and segmental optimization. Despite these advances, existing RL-based approaches often lack fine-grained feedback on error types and distributions, limiting their ability to address recurring issues. Moreover, their corrective mechanisms are typically weak, offering limited and untimely guidance. Finally, the limited diversity in benchmark datasets restricts model generalization to unseen scenarios \cite{RLTF:23}.

\section{Problem Setting}
We aim to enhance automated code generation capabilities of Large Language Models (LLMs) by addressing accuracy, diversity, error correction, code quality, and adaptability. Formally, given a high-level specification \(D\), the task is to generate a sequence of code tokens \(W=\{w_1,w_2,\ldots,w_T\}\) that maximizes \(P(W|D,\theta)\), with the objective \(\theta^*=\arg\max_{\theta}\mathbb{E}_{D\sim\mathcal{D}}[\log P(W|D,\theta)]\). \textbf{Accuracy} requires minimizing syntactical or logical errors, while limited \textbf{diversity} in input-output pairs \(\mathcal{X}\) restricts generalization. Efficient \textbf{error correction} is crucial for identifying and fixing errors in \(W\). \textbf{Code quality} involves adhering to standards, style guidelines, and managing complexity \(C(W)\). \textbf{Adaptability} depends on the model’s capacity to incorporate feedback for continuous improvement, which is constrained without robust memory. The \textbf{FALCON} framework leverages a long-term memory buffer \(\mathcal{M}_{\text{long}}=\{(D_i,W_i,T_i,F_i)\}_{i=1}^N\) and a short-term memory buffer \(\mathcal{M}_{\text{short}}=\{(D_j,W_j,T_j,F_j)\}_{j=1}^M\) to utilize diverse feedback for fine-tuning \(\theta\). FALCON optimizes \(\theta\) by maximizing a composite reward \(R(W,F)=\alpha T(W)+\beta S(W)+\gamma C(W)+\delta E(W)\), where \(T(W)\), \(S(W)\), \(C(W)\), and \(E(W)\) denote unit test, style, complexity, and error feedback, respectively, yielding \(\theta^*=\arg\max_{\theta}\mathbb{E}_{D\sim\mathcal{D}}[\mathbb{E}_{W\sim P(W|D,\theta)}[R(W,F)]]\). We assume exchangeability of \(\mathcal{D}\), independence of feedback \(F\) given \(W\), sufficient memory capacity in \(\mathcal{M}_{\text{long}}\) and \(\mathcal{M}_{\text{short}}\), and effective feedback mechanisms. Given tasks \(\mathcal{T}=\{D_i\}_{i=1}^N\), the ultimate goal is \(\theta^*=\arg\max_{\theta}\frac{1}{N}\sum_{i=1}^N R(W_i,F_i)\) with \(W_i\sim P(W|D_i,\theta)\), ensuring continuous improvement by leveraging both historical and recent feedback to enhance code generation quality.
\section{Methodology}
The FALCON framework enhances code generation by LLMs through unit testing and reinforcement learning, leveraging long-term and short-term memory buffers. Task descriptions, generated code, and feedback (e.g., compilation results, code style, complexity) are stored in the long-term memory buffer, allowing the model to reference high-quality code, avoid errors, and adhere to standards. A judge model evaluates the code, calculates feedback-based rewards, and updates model parameters via reinforcement learning. This dual memory mechanism enables the model to improve code quality using historical data while adapting to new tasks. The method section includes three components: Meta-Reinforcement Learning with Dual Memory Buffers~\ref{sec : 2}, introducing the dual memory structure; and Long-Term Memory Feedback~\ref{sec : 3}, optimizing with historical data. See Figure~\ref{fig:system_architecture}.
\subsection{MRLF: Meta-Reinforcement Learning with Dual Memory Buffers}\label{sec : 2}
We propose the MRLF algorithm~\ref{alg:mrlf_algorithm}, which involves random task sampling from the task distribution during training. Previous works have indicated that different data sampling strategies have varying impacts on model information extraction \cite{MSL:24}. Consequently, we implement both long and short memory sequences. The long memory strategy stores the solutions generated for each problem and the compiler feedback results, whereas the short memory sequence selects the latest samples and unit test feedback from each current iteration. To address repetitive runtime and compilation errors, the long memory strategy categorizes and stores various errors by their types and records the corresponding error lines, enabling fine-grained reward allocation.
\begin{algorithm}[htbp]
\caption{MRLF Algorithm}
\label{alg:mrlf_algorithm} 
\begin{algorithmic}
\Require Task distribution $\mathcal{T}$
\Ensure Updated model parameters $\theta$
\State Initialize $LMB$, $SMB$, $\theta$
\State Populate $LMB$, $SMB$ via few-shot demonstrations 
\Repeat
    \State Sample batch $\{T_i\}$ from $\mathcal{T}$
    \For{each task $T_i$}
        \State \textbf{Code Generation:} Generate code $\hat{W}$ and test results; record in $SMB$
        \State \textbf{Adaptation:} Set $\theta_i = \theta$; sample mini-batch from $LMB$, $SMB$
        \State Compute inner loss: $L_{\text{inner}} = \sum L_j$ $(j \in \{\text{sl, coarse, error, complexity, style, negative}\})$
        \State Update $\theta_i$: $\theta_i \leftarrow \theta_i - \alpha \nabla_{\theta_i} L_{\text{inner}}(\theta_i)$
        \State \textbf{Evaluation:} Assess $\theta_i$ on $T_i$; document outcomes
    \EndFor
    \State \textbf{Meta-Update:} $L_{\text{meta}} = \sum_{i=1}^{N} L_{\text{inner}}(\theta_i)$; refine $\theta$: $\theta \leftarrow \theta - \beta \nabla_\theta L_{\text{meta}}$
\Until{Convergence}
\State \Return $\theta$
\end{algorithmic}
\end{algorithm} We initialize two memory buffers (LMB, SMB) and model parameters \( \theta \), then randomly sample tasks from the task distribution, executing steps for each one. First, the corresponding code is generated and tested, with results stored in SMB. During adaptation, the algorithm extracts experimental data from LMB and SMB to aid rapid adaptation to the current task. The inner loss function \( L_{\text{inner}} \), incorporating factors like accuracy, complexity, style, and negative examples, is optimized using policy gradient. In evaluation, the optimized parameters are tested, and results are stored for future use. Finally, the losses from all tasks are aggregated to compute the meta-loss \( L_{\text{meta}} \), used to update global parameters \( \theta \). Algorithm~\ref{alg:mrlf_algorithm} summarizes the framework. For inner loop optimization, our method explores the target space by combining unit test feedback, code complexity, and style norms, using the generated code \( \hat{w} \) to construct the reinforcement learning loss function as shown below:
\begin{equation}
L_{rl} = -\sum_{t=S_{\text{fine}}}^{E_{\text{fine}}} R_{\text{fine}}(\hat{w}_t) \log p(\hat{w}_t | D, \theta, \hat{w}_{1:t-1})
\label{eq:rl_loss}
\end{equation}
where \(R_{\text{fine}}(\ast)\) represents the reward coefficient, and \(S_{\text{fine}}\) and \(E_{\text{fine}}\) denote the start and end positions of the code snippet, respectively. These values are determined based on different types of feedback. To stabilize the training process, we adopt the supervised learning loss \( L_{sl} \) by minimizing the cross-entropy loss as shown below:
\begin{equation}
L_{sl} = -\log P(w \mid D, \theta) = - \sum_{t=1}^{T} \log p(w_t \mid D, \theta, w_{1:t-1})
\end{equation}

\begin{figure}[htbp]
    \centering
    \includegraphics[width=\linewidth]{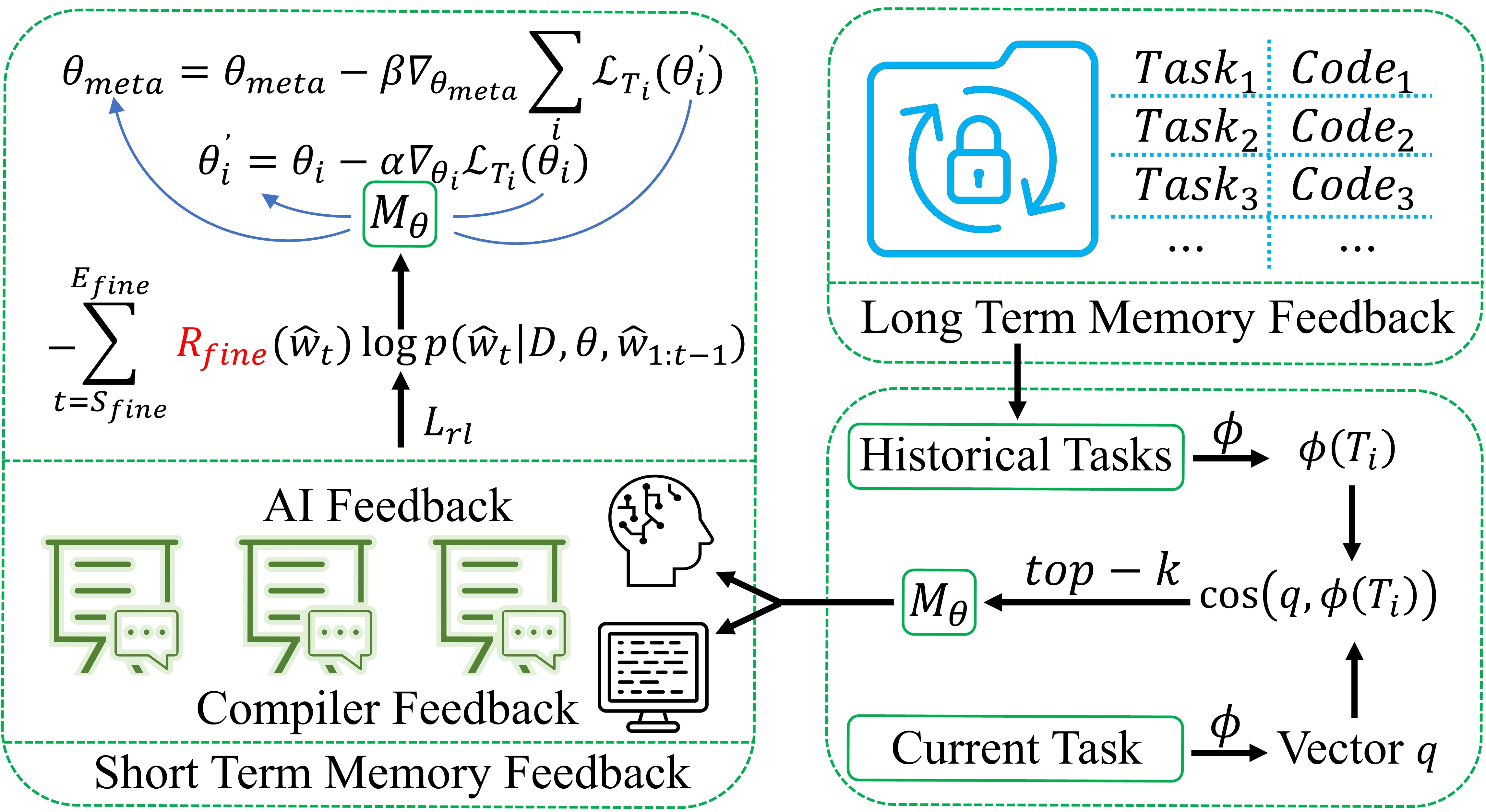}
    \caption{The framework integrates meta-reinforcement learning with both long-term and short-term memory feedback. Long-term memory retrieves historical tasks, while short-term memory provides AI and compiler feedback to refine code generation.}
    \label{fig:meta_framework}
\end{figure}

As depicted in Figure~\ref{fig:meta_framework}, we employ a meta-reinforcement learning framework to optimize code generation by integrating both long- and short-term memories for enhanced adaptability. From a global perspective, Long-Term Memory \( D_{\text{long}} \) stores historical tasks, generated codes, and feedback to provide valuable context. From a local perspective, Short-Term Memory \( D_{\text{short}} \) focuses on recent feedback to enable real-time adjustments. This approach leverages the MAML framework \cite{finn2017modelagnosticmetalearningfastadaptation} for efficient task adaptation with minimal updates.\\

\textbf{Short-Term Memory Adaptation.}  
Short-Term Memory is utilized to adapt the model locally by adjusting its parameters based on recent feedback. For each task \( \mathcal{T}_i \), the inner loop optimization updates the parameters:
\begin{equation}
\theta_i' = \theta_i - \alpha \nabla_{\theta_i} \mathcal{L}_{\mathcal{T}_i} (\theta_i)
\end{equation}
where \( \alpha \) is the learning rate and \( \mathcal{L}_{\mathcal{T}_i} \) is the task-specific loss function.\\
\textbf{Global Optimization.}  
The outer loop performs global optimization of the meta-learning parameters \( \theta_{\text{meta}} \) by aggregating feedback across multiple tasks:
\begin{equation}
\theta_{\text{meta}} = \theta_{\text{meta}} - \beta \nabla_{\theta_{\text{meta}}} \sum_{i} \mathcal{L}_{\mathcal{T}_i} (\theta_i')
\end{equation}
where \( \beta \) is the meta-learning rate. This ensures better generalization across tasks.\\
\textbf{Final Optimization.}  
The overall framework combines short- and long-term memory feedbacks with meta-reinforcement learning to achieve coordinated optimization for both global generalization and local task adaptation:
\begin{equation}
\theta_{\text{final}} = \text{Optimize}\left(\theta_{\text{meta}}, \theta, \{\theta_i'\}\right)
\end{equation}

\begin{table*}[!t]
\centering
\setlength{\tabcolsep}{10pt}
\caption{Quantitative evaluation on the APPS benchmark. “Intro”: introductory, “Inter”: interview, “Comp”: competition-level tasks. To ensure a fair comparison, we apply these RL-based methods, including PPOCoder, CodeRL, and RLTF, using the same base model, CodeT5, as a backbone. We also compare with models that have a larger number of parameters.}
\renewcommand{\arraystretch}{1.2}
\arrayrulecolor{black}
\arrayrulewidth=1pt
\resizebox{\textwidth}{!}{
\begin{tabular}{lcccccccccc}
\toprule
\multirow{2}{*}{\textbf{Method}} & \multirow{2}{*}{\textbf{Size}} & \multicolumn{4}{c}{\textbf{pass@1}} & \multicolumn{4}{c}{\textbf{pass@5}} \\ 
\cmidrule(lr){3-6} \cmidrule(lr){7-10}
& & \textbf{Intro} & \textbf{Inter} & \textbf{Comp} & \textbf{All} & \textbf{Intro} & \textbf{Inter} & \textbf{Comp} & \textbf{All} \\ \midrule
Codex & 12B & 4.14 & 0.14 & 0.02 & 0.92 & 9.65 & 0.51 & 0.09 & 2.25 \\
GPT-Neo & 2.7B & 3.90 & 0.57 & 0 & 1.12 & 5.50 & 0.80 & 0 & 1.58 \\
CodeT5 base & 770M & 3.85 & 0.58 & 0.02 & 1.05 & 8.52 & 1.53 & 0.25 & 2.82 \\
PPOCoder & 770M & 4.06 & 0.79 & 0.15 & 1.32 & 9.97 & 2.06 & 0.70 & 3.37 \\
CodeRL & 770M & 7.08 & 1.86 & 0.75 & 2.69 & 16.37 & 4.95 & 2.84 & 6.81 \\
RLTF & 770M & 8.40 & 2.28 & 1.10 & 3.27 & 18.60 & 5.57 & \textbf{3.70} & 7.87 \\ \midrule
\rowcolor{gray!30}
\textbf{Ours} & 770M & \textbf{8.60} & \textbf{2.56} & \textbf{1.24} & \textbf{3.50} & \textbf{19.75} & \textbf{5.85} & 3.57 & \textbf{8.17} \\
\bottomrule
\end{tabular}
}
\label{table1}
\end{table*}
\subsection{Long-Term Memory Feedback}\label{sec : 3}
\label{longterm}
Retrieving information from long-term memory helps improve code quality. We use the FAISS framework \cite{douze2024faisslibrary} to retrieve relevant historical code, feedback, and evaluation scores. Task descriptions and feedback are transformed into embedding vectors and then indexed. During code generation, a query vector from the current task retrieves the top-k most similar historical data to guide the process and avoid past errors. The prompt template is provided in the appendix. Consider a set of historical data $\mathcal{D} = {(t_i, f_i, e_i)}_{i=1}^{n}$, where $t_i$ represents the task description, $f_i$ is the corresponding feedback, and $e_i$ is the evaluation score. We use an embedding function $\phi(\cdot)$ to transform these tasks and feedback into embedding vectors $\boldsymbol{v}_i = \phi(t_i, f_i)$ and index them with FAISS.
During the code generation phase, the current task description $\boldsymbol{t}_{\text{current}}$ and feedback $\boldsymbol{f}$ are transformed into a query vector $\boldsymbol{q} = \phi(\boldsymbol{t}_{\text{current}})$. We compute the similarity between the query vector $\boldsymbol{q}$ and the historical vectors $\boldsymbol{v}_i$ using cosine similarity $\text{cos}(\boldsymbol{q}, \boldsymbol{v}_i)$, and retrieve the top-$k$ most similar historical tasks. The retrieval process can be represented as:
\begin{equation}
\{(t_{i_1}, f_{i_1}, e_{i_1}), \dots, (t_{i_k}, f_{i_k}, e_{i_k})\} = \text{Top-}k \left( \boldsymbol{v}_i \mid i = 1, 2, \dots, n \right)
\end{equation}
By referencing these most relevant historical tasks and feedbacks, the system can guide the current code generation process with past mistakes avoided and ultimate code quality improved.
\subsection{Short-Term Memory Feedback}
During the reinforcement learning phase, we utilize the generated code \( \hat{w} \) to optimize the policy based on the reinforcement learning loss defined in Equation~(\ref{eq:rl_loss}). As previously described, this loss incorporates reward signals such as unit test feedback, code complexity, and style norms, and is computed over the token span from \(S_{\text{fine}}\) to \(E_{\text{fine}}\).

\textbf{Compiler Feedback.} For compiler feedback, we adopt the same settings as CodeRL:
\begin{equation}
R_{\text{coarse}}(\hat{W}) = 
\left\{
\begin{array}{ll}
 \phantom{-}1.0, & \text{if } FB(\hat{W}) \text{ is pass} \\
-0.3, & \text{if } FB(\hat{W}) \text{ is failure} \\
-0.6, & \text{if } FB(\hat{W}) \text{ is runtime error} \\
-1.0, & \text{if } FB(\hat{W}) \text{ is syntax error}
\end{array}
\right.
\end{equation}
\[
S_{\text{coarse}} = 0, \quad E_{\text{coarse}} = T
\]
where \(R_{\text{coarse}}\) is based on compiler feedback with the start and end positions set to 0 and \(T\).\\
\textbf{Adaptive Feedback.}
To enhance the model's efficiency in handling various programming tasks, we devise a mechanism that dynamically adjusts rewards based on the proportion of passes to failures in unit tests. This strategy encourages the model not only to pass unit tests but also to learn from failures, thereby improving its problem-solving capabilities. The reward is calculated as:
\begin{equation}
R_{\text{error}}(\hat{W}) = -0.3 + 1.3 \times \frac{N_{\text{pass}}}{N_{\text{pass}} + N_{\text{fail}}}
\end{equation}
\textbf{Coding Style Feedback.}
To further enhance the quality of the generated code, we employ AI Feedback to optimize coding style. An evaluation model scores the generated code based on adherence to the expected coding style standards. The scoring system ranges from -1 to 2, and these evaluation scores are directly used as reward signals in the reinforcement learning process to guide the model toward producing higher-quality code. \\
\textbf{Complexity Feedback.} Just like with coding style, we use AI Feedback to evaluate complexity and calculate rewards based on the scores.\\
\noindent\textbf{Long-term Error Type Feedback.} A reward mechanism combining short-term error recall, current test errors, and long-term memory of past performance dynamically adjusts rewards. The formula is:  
\begin{equation}  
R_{\text{negative}} = -\sum_{\text{error}} N_{\text{error}} \times P_{\text{error}}  
\end{equation}  
where \( N_{\text{error}} \) represents short-term feedback on current task errors, and \( P_{\text{error}} \) represents long-term error proportions. This mechanism reduces errors and enhances the accuracy and quality of code generation.

\section{Experiment}
\subsection{Quantitative Evaluation on APPS}
To ensure a fair comparison, we use the CodeT5 770M model as our baseline. Our benchmarks include the latest advancements that integrate reinforcement learning (RL) with large language models (LLMs), particularly CodeRL, PPOCoder, and RLTF. For evaluation, we apply the same benchmarks and settings used in these previous works. As shown in Table~\ref{table1} for the experimental results, our FALCON approach delivers additional performance improvements and surpasses other RL-based methods, indicating that RL with appropriate feedback can effectively improve the model output space and thereby enhance the quality of code generation. In particular, our method achieves the highest pass@1 rates of 8.60\%, 2.56\%, and 1.25\% in the Introductory, Interview, and Competition categories, respectively.

\subsection{Quantitative Evaluation on HumanEval and MBPP}

To further validate the effectiveness of our method, we evaluate the zero-shot performance of the DeepSeek-Coder-Instruct model, trained with our method on our custom dataset, using the well-established MBPP and HumanEval benchmarks. We also compare these results against other reinforcement learning methods, such as PPOCoder and RLTF. The experimental results are illustrated in Table~\ref{table4}.  

\begin{table}[ht]
\centering
\setlength{\abovecaptionskip}{5pt} 
\setlength{\belowcaptionskip}{8pt} 
\caption{The results of pass@1 on the MBPP and HumanEval benchmarks.}
\resizebox{0.8\linewidth}{!}{ 
\begin{tabular}{c | c c}
\toprule
\textbf{Model} & \textbf{Humaneval} & \textbf{MBPP} \\
\midrule
DeepSeek-Coder-Instruct & 73.8 & 74.9  \\
PPOCoder & 76.8 & 76.2 \\
RLTF & 76.8 & 75.9 \\
\midrule
\rowcolor{gray!30}
\textbf{Ours} & \textbf{82.9} & \textbf{80.7} \\
\bottomrule
\end{tabular}
}
\label{table4}
\end{table}
Compared to other reinforcement learning methods, our method consistently achieves the best performance on both the HumanEval and MBPP benchmarks. The significant advantage of our method can be attributed to its diversified feedback mechanism. Unlike other methods that may focus on a single metric, our method continuously optimizes the model's generation capability through multi-dimensional feedback. This approach demonstrates a strong ability to enhance the generation of correct code and proves particularly effective in complex tasks.
\subsection{Quantitative Evaluation on CODAL-Bench}
\begin{figure}[htbp]
\centering
\includegraphics[width=\linewidth]{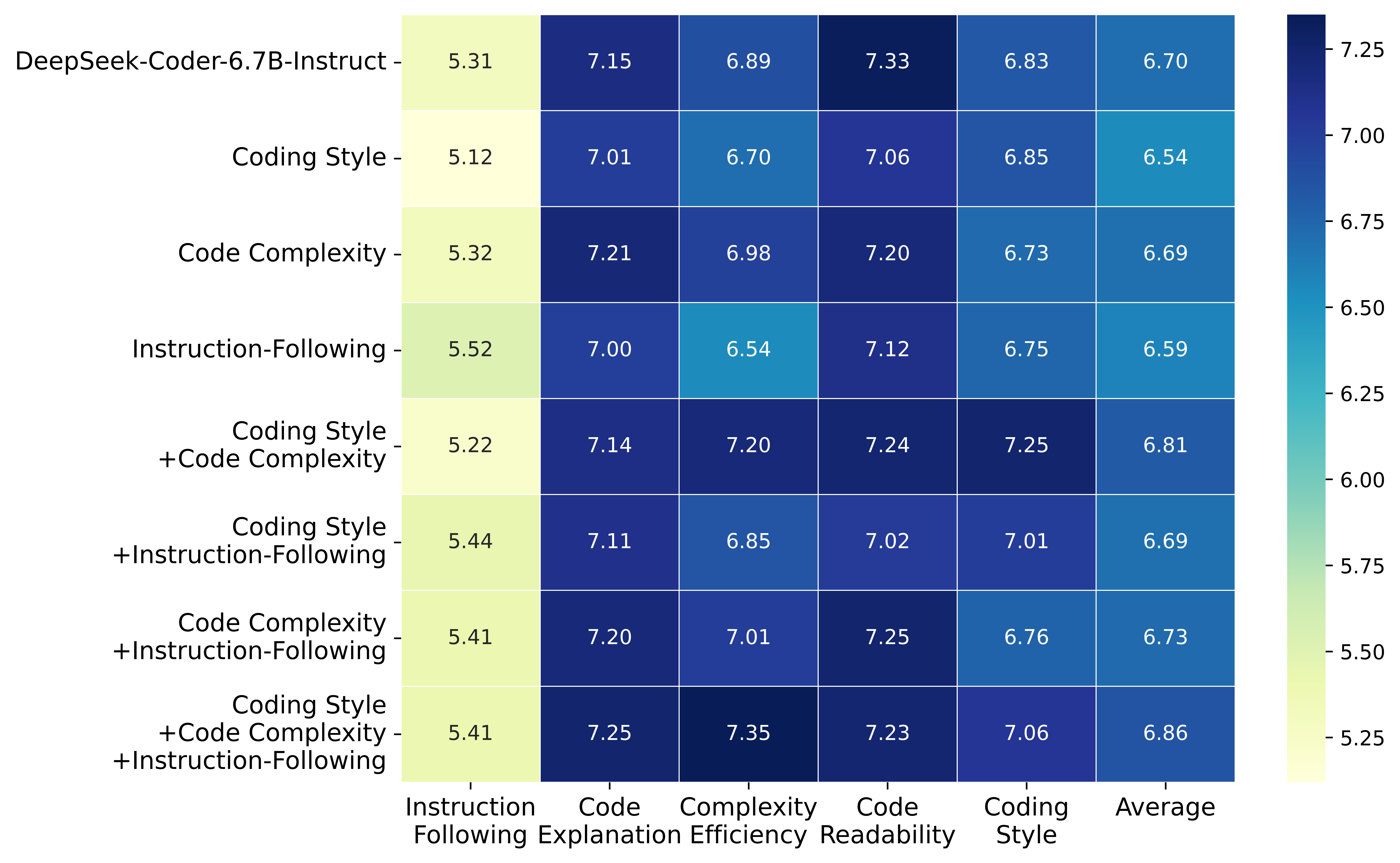}
\caption{Quantitative evaluation on CODAL-Bench}
\label{heatmap.png}
\end{figure}

In addition to evaluating the functional correctness of the code, we adopt CODAL-Bench, a rigorous and comprehensive benchmark for LLM consistency in coding preferences to validate the effectiveness of short-term memory feedback. DeepSeek-Coder-Instruct-6.7B model is used and the results are illustrated in Figure~\ref{heatmap.png}. It is found that there is a noticeable improvement in various coding preferences, particularly in Code Complexity and Coding Style after implementing the FALCON framework. This observation is attributed to the inclusion of feedback on these aspects in short-term memory. However, the improvement in Instruction Following is not as significant.

\subsection{Ablation Studies}
\noindent\textbf{The Influence of Models.} To validate the scalability and robustness of our framework, we conduct experiments with the larger model, DeepSeek-Coder-Instruct-6.7B, to further evaluate its performance. Notably, the improvements in introductory-level tasks are significant, which can be attributed to the use of long-term memory that enhances the quality of generated data and further unlocks the model's potential. The results are illustrated in Table~\ref{impact of model}.

\begin{table}[htbp]
\caption{Different large language models as the backbone}
\label{impact of model}
\centering
\begin{tabular}{c c c | c c c c}
\toprule
\textbf{Model} & \textbf{Size} & \textbf{Method} & \textbf{Intro} & \textbf{Inter} & \textbf{Comp} & \textbf{All} \\
\midrule
CodeT5 & 770M & w/o & 3.85 & 0.58 & 0.02 &  1.12\\
\rowcolor{gray!30}
CodeT5 & 770M & w & 8.60 & 2.56 & 1.25 & 3.50 \\
DeepSeek-Coder & 6.7B & w/o & 16.70 & 7.20 & 2.30 & 8.12 \\
\rowcolor{gray!30}
DeepSeek-Coder & 6.7B & w & 22.40 & 8.52 & 3.70 & 10.33 \\
\bottomrule
\end{tabular}
\end{table}


\noindent\textbf{The Influence of Memory.} To validate the impact of long-term and short-term memories on code generation capabilities, we conduct ablation experiments using CodeT5 as the base model and test it on the APPS dataset. As shown in Table~\ref{Impact of Memory}, the experimental results indicate that both long- and short-term memory feedback enhances the model's code generation performance effectively, while the short-term memory feedback demonstrates a more significant improvement. This improvement can be attributed to the effective reward design which plays a positive role in fine-tuning the model. 

\begin{table}[ht]
\caption{Effect of long and short memories on different performance metrics}
\label{Impact of Memory}
\centering
\setlength{\tabcolsep}{6pt} 
\begin{tabular}{c c | c c c c}
\toprule
\textbf{Long Memory} & \textbf{Short Memory} & \textbf{Intro} & \textbf{Inter} & \textbf{Comp} & \textbf{All} \\
\midrule
- & - & 3.85 & 0.58 & 0.02 & 1.12 \\
\checkmark & - & 4.14 & 0.74 & 0.02 & 1.28 \\
- & \checkmark & 7.20 & 1.86 & 0.70 & 2.70 \\
\rowcolor{gray!30}
\checkmark & \checkmark & 8.60 & 2.56 & 1.25 & 3.50 \\
\bottomrule
\end{tabular}
\end{table}
\section{Conclusions}
In this work, we propose \textbf{FALCON}, a novel RL-based feedback~\cite{he2024enhancing} framework that enhances automated code generation by integrating long-term and short-term memory feedbacks within a meta-reinforcement learning strategy to better capture human intent like RL based preference learning. Long-term memory retains past interactions to improve code quality and reduce repetitive mistakes, while short-term memory enables immediate adjustments based on recent feedback from compilers and AI systems. This dual-memory approach addresses limitations in existing models that struggle to align code generation with user intent, especially in specialized tasks or edge cases. By incorporating non-differentiable code features like style and complexity into the feedback loop, FALCON ensures that the generated code is not only functionally correct but also adheres to real-world programming standards. Extensive evaluations on benchmarks including APPS, HumanEval, and CODAL-Bench demonstrate that FALCON outperforms existing RL-based methods, achieving higher functional correctness and better coding style adherence.
\bibliographystyle{IEEEbib}
\bibliography{icme2025references}
\end{document}